\documentclass{article}
\usepackage{spconf}

\usepackage{epsfig}
\usepackage{enumerate}
\usepackage[fleqn]{amsmath}
\usepackage{amsthm}
\usepackage{amscd}
\usepackage{amssymb}
\usepackage{subfigure}
\usepackage{lscape}
\usepackage{fancyhdr}
\usepackage{graphics} 
\usepackage{epsfig} 
\usepackage{mathptmx} 
\usepackage{times} 
\usepackage{graphicx}
\usepackage{url}
\usepackage{pifont}
\usepackage{hhline}
\usepackage{multirow}
\usepackage{rotating}
\usepackage{amsbsy}
\usepackage{array}
\usepackage[linesnumbered,lined,boxed,ruled,commentsnumbered]{algorithm2e}
\usepackage[T1]{fontenc}

\usepackage{amsmath,mathtools}
\newcommand{\sunderb}[2]{
	\mathclap{\underbrace{\makebox[#1]{$\cdots$}}_{#2}}
}

\DeclareMathAlphabet{\mathcal}{OMS}{cmsy}{m}{n}
\title{An LBP-HOG Descriptor Based on Matrix Projection For Mammogram Classification}
\twoauthors
  {Zainab Alhakeem  }
	{School of Electrical and Electronic Engineering, \\ Yonsei University \\
	Email: zrkhakim@dsp.yonsei.ac.kr}
   {{\hspace{-1.5cm}} Se-In Jang\sthanks{Corresponding Author.}  }
	{{\hspace{-1.5cm}} Department of Statistics and Applied Probability, \\ {\hspace{-1.5cm}}National University of Singapore \\ {\hspace{-1.5cm}}{  Email: sijang@nus.edu.sg}}

\begin{document}
%
\maketitle
\begin{abstract}
In image based feature descriptor design, 
local information from image patches are extracted using iterative scanning operations 
which cause high computational costs.
In order to avoid such scanning operations, we present  matrix multiplication based local feature descriptors, namely a Matrix projection based Local Binary Pattern (M-LBP) descriptor and a Matrix projection based Histogram of Oriented Gradients (M-HOG) descriptor.
Additionally, an integrated formulation of M-LBP and M-HOG (M-LBP-HOG) is also proposed to perform the two descriptors together in a single step.
The proposed descriptors are evaluated using a publicly available mammogram database. The results show promising performances in terms of classification accuracy and computational efficiency.
\end{abstract}
\section{Introduction}
\label{sec:intro}

Breast cancer is the most common and leading cause of cancer death in women \cite{bray2018global}. 
For early diagnosis of the breast cancer, mammography X-rays are manually examined by specialized doctors. 
In order to automate the diagnosis,  Computer-Aided Diagnosis (CAD) systems  \cite{soltanian2001shape, bekker2016multi, oliver2007classifying, shastri2018density} have been developed for mammogram classification between benign and malignant. 
Such mammogram CAD systems depend on precise segmentations of the target tumors (i.e., complex lesion boundary of the tumors) to extract reliable features from its shape information \cite{soltanian2001shape, bekker2016multi}. 
To avoid this dependency on the segmentation problem, local feature descriptors have been frequently utilized such as 
the Local Binary Pattern (LBP) method for texture information \cite{ojala1994performance} and the Histogram of Oriented Gradients (HOG) method for shape information \cite{dalal2005histograms}.

Specifically, the LBP descriptor extracts illumination invariant image features which can compensate for the illumination variation through the binary patterns \cite{liu2017local}.
The HOG descriptor builds geometric and  photometric invariant image features obtained from local gradient histogram of horizontal and vertical directions \cite{dalal2005histograms}. 
As a combination of LBP and HOG features, an LBP-HOG method takes both useful properties  
and outperforms the utilization of each descriptor solely in \cite{wang2009hog, konstantinidis2016building, zhang2011boosted}. 
However, such combinations of local descriptors should be separately performed one after the other.
Moreover, when faced with the use of local descriptors, iterative scanning operations for local image regions are a heavy burden due to their long time-consuming process.
In order to avoid such scanning operations, 
in \cite{liu2019pedestrian}, a Difference Matrix Projection (DMP) based descriptor was proposed to take local representation similar to HOG using one step matrix multiplication. 
However, the DMP descriptor has partially completed the HOG descriptor without forming a matrix based block normalization.

Our contributions of this work are as follows: 1) A new Matrix formulation of LBP (M-LBP) based on the matrix projection to exclude the iterative scanning; 2) A new Matrix formulation of HOG (M-HOG) based on the DMP descriptor; and 3)  An integrated formulation of M-LBP and M-HOG (M-LBP-HOG) to avoid performing each descriptor separately.
%

The remainder of the paper is organized as follows: Section \ref{sec:prelim} includes a brief review of the LBP descriptor and the matrix based pixel difference computation. In Section \ref{sec:CF}, we propose two image descriptors in matrix formulations. Then we construct an integrated formulation of these two descriptors. Section \ref{sec:exps} presents our experiments of the proposed method using a public mammogram database. Concluding remarks are given in Section \ref{sec:conc}.

\section{Preliminaries}
\label{sec:prelim}

\subsection{Local Binary Pattern (LBP)}
\label{ssec:LBC}


The Local Binary Pattern (LBP) \cite{ojala1994performance} is a simple and popular descriptor adopted in various applications (e.g., face and palm print recognitions \cite{ahonen2006face, sepas2017light,han2009palmprint}). 
The LBP takes an iterative loop using an odd size window (e.g., $3 \times 3$ window) for image scanning to describe each central pixel with its local neighborhood pixels ($P=8$). The LBP can be expressed as:
\begin{equation} \label{eq.lbp}
\footnotesize
{LB{P_{P}} = \sum\limits_{i = 0}^{P - 1} {f(d_i){2^i}} },
\end{equation}
\noindent where $d_i = {s_i} - {x_c}$ is the pixel difference between the neighborhood pixel ${s_i}$ and the center pixel  ${x_c}$ for each direction from ${{0}^{{}^\circ }}$ to ${{315}^{{}^\circ }}$. $f(\cdot)$ is the thresholding operation given by 
$f\left( x \right) = \left\{ \hspace{-0.3cm}{\begin{array}{*{20}{c}}
	{\begin{array}{*{20}{c}}
		{1,}& \hspace{-0.2cm} \text{if} \hspace{0.2cm} {x \ge 0}
		\end{array}}\\
	{\begin{array}{*{20}{c}}
		{0,}& \hspace{-0.2cm} \text{if} \hspace{0.2cm} {x < 0}
		\end{array}}
	\end{array}}\hspace{-0.3cm}  \right.$. 

 \subsection{A Matrix based Pixel Difference Computation}
\label{ssec:DMP}

In feature descriptors (e.g., LBP and HOG), an essential step is the computation of the pixel difference within each local window. However, the window based technique suffers from the iterative computation for each pixel. 
In \cite{liu2019pedestrian}, a matrix based pixel difference computation is formulated to globally calculate the pixel differences based on pre-calculated projection matrices:
\begin{equation} \label{eq.HV} 
\footnotesize
\begin{array}{l}
{{\bf{H}}_l} = \left. {\left( {\begin{array}{*{20}{c}}
		0&{\bf{I}}\\
		{\bf{I}}&0
		\end{array}} \right)} \right\}\begin{array}{*{20}{c}}
{N - l}\\
l
\end{array}\in \mathbb{R}{^{N \times N}},\\
{{\bf{V}}_l} = \left. {\left( {\begin{array}{*{20}{c}}
		0&{\bf{I}}\\
		{\bf{I}}&0
		\end{array}} \right)} \right\}\begin{array}{*{20}{c}}
{M - l}\\
l
\end{array} \in \mathbb{R}{^{M \times M}},
\end{array}
\end{equation}

\noindent where ${\bf{H}}_l$ and ${\bf{V}}_l$ are matrices to horizontally and vertically compute image derivatives respectively. ${\bf{I}}$ is the identity matrix, $M$ and $N$ respectively are the number of rows and columns of the input image ${\bf{X}} \in {\mathbb{R}^{M \times N}}$, and $l$ indicates a shifting distance between two neighbor pixels. For example, when $l = 1$, 
hortizontally and vertically one-pixel difference matrices are given by ${\bf{XH}}_1^{} - {\bf{X}}$ and ${\bf{V}}_1^{}{\bf{X}} - {\bf{X}}$ respectively.

\section{An LBP-HOG Descriptor Based on Matrix Multiplication}
\label{sec:CF}


First, we propose a Matrix based LBP descriptor (M-LBP). Additionally, we present a complete version of Matrix based HOG descriptor (M-HOG). We then propose a M-LBP-HOG descriptor which enjoys the properties of M-LBP and M-HOG in a single step.

 
\subsection{A Matrix Based LBP Descriptor}
\label{ssec:PropLBP}

In \cite{ojala1994performance}, the LBP computes the pixel differences for each window in eight directions. 
The proposed Matrix based LBP descriptor (M-LBP) can be formulated as the weighted sum of eight directional difference matrices as follows:  
\begin{equation} \label{eq.PropLBP}
f^{LBP}\left( {\bf{X}} \right) = {{\bf{Z}}^{LBP}  = \sum\limits_{i = 1}^{P-1} {\sigma ({{\bf{D}}_{i}})} {2^{i}} },
\end{equation} 
\noindent where 
$\sigma \left(  \cdot  \right)$ is the step function, 
${{\bf{D}}_{i}^{} = {{\bf{S}}_i} - {\bf{X}}}$, and
\begin{equation} \label{eq.PropSLBP}
{\begin{array}{*{20}{l}}
{{\bf{S}}_{1}^{} = {\bf{V}}_1^{ - 1}{\bf{XH}}_1^{},}	&{{\bf{S}}_{5}^{} = {\bf{V}}_1^{}{\bf{XH}}_1^{ - 1},}\\
{{\bf{S}}_{2}^{} = {\bf{V}}_1^{ - 1}{\bf{X}},}				&{{\bf{S}}_{6}^{} = {\bf{V}}_1^{}{\bf{X}},}\\
{{{\bf{S}}_{3}} = {\bf{V}}_1^{ - 1}{\bf{XH}}_1^{ - 1},}	&{{\bf{S}}_{7}^{} = {\bf{V}}_1^{}{\bf{XH}}_1^{},}\\
{{\bf{S}}_{4}^{} = {\bf{XH}}_1^{ - 1},}					&{{\bf{S}}_{8}^{} = {\bf{XH}}_1^{},}
\end{array}}
\end{equation}
\noindent are shifting matrices for the eight directions using 
${\bf{V}}_{l}^{}$ and ${\bf{H}}_{l}^{}$ at $l=1$ in equation \eqref{eq.HV}.
Therefore, in equation \eqref{eq.PropLBP}, the weighted sum   provides the pixel differences for each matrix in eight directions.

\begin{figure}[h]

	\hspace{-1.3cm} 
	\includegraphics[width=1.3\linewidth]{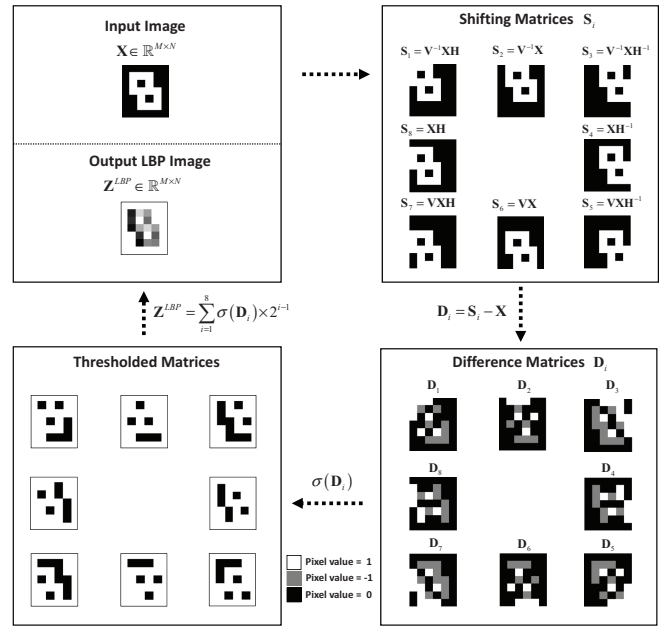}   
	\caption{An overview of the proposed M-LBP descriptor using a synthetic image.} %
	\label{fig.lbp_prop} 
\end{figure}

Based on a synthetic binary image, 
the process of our proposed M-LBP descriptor is illustrated in Fig.\ref{fig.lbp_prop}.
The shifting matrix ${\bf{S}}_i$ is the shifting matrix created by the multiplication operation between the pre-calculated projection matrices with the input image.
The difference matrix ${\bf{D}}_i$ is calculated by subtracting the input image ${\bf{X}}$ from the shifting matrices. 
The thresholding operation can be easily obtained using a step function.
The output M-LBP image is then given by the sum of the thresholded matrices multiplied by the eight binary codes.

\vspace{-0.2cm}

\subsection{A Matrix Based HOG Descriptor}
\label{ssec:PropHOG}
In \cite{liu2019pedestrian}, as an approximated HOG descriptor, a Difference Matrix Projection (DMP) descriptor in matrix form was proposed. Compared to HOG, the DMP descriptor uses the first and the second order gradient information together. This DMP descriptor consists of 1) the matrix based pixel difference computation shown in Section \ref{ssec:DMP}, 2) a non-overlapping matrix based average pooling and 3) an iterative block normalization which is not in matrix form. 

In the proposed M-HOG, an overlapping matrix based average pooling and a matrix based block normalization are constructed using predefined projection matrices 
\begin{equation} \label{eq.overlapPooling_L}
	\tiny
	{\bf{L}}_{c,v} = 
	\underbrace {
		\begin{bmatrix}
			1\text{ } & \sunderb{4.6em}{1\times c} & \hspace{-0.15cm} 1\text{ } &  & &&0\\
			&  &  { \hspace{-0.2cm} \overbrace{1\text{}}^{v} }& \sunderb{6.8em}{1\times c} & \hspace{-0.1cm} 1 &  &\\ 
			0 &  &  &  { \ddots}& { \hspace{-0.1cm} \overbrace{\text{}1}^{v} } &\sunderb{4.6em}{1\times c}&1
	\end{bmatrix}  }_M  
\end{equation} and
\begin{equation} \label{eq.overlapPooling_R}
	\tiny
	{{\bf{R}}_{c,v} = 
		\underbrace { 
			\begin{bmatrix}
				1\text{ } & \sunderb{4.6em}{1\times c} & \hspace{-0.15cm} 1\text{ } &  & &&0\\
				&  &  { \hspace{-0.2cm}\overbrace{1\text{}}^{v} }& \sunderb{6.8em}{1\times c} & \hspace{-0.1cm} 1 &  &\\ 
				0 &  &  &  { \ddots}& { \hspace{-0.1cm}\overbrace{\text{}1}^{v} } &\sunderb{4.6em}{1\times c}&1
		\end{bmatrix}}_N} ^T ,
\end{equation}
where the subscript $c$ indicates a variable for the desired local window size (i.e., $2 \times 2 $ window at $c=2$), and
the subscript $v$ is a variable for the overlapping size between two windows (i.e., $0<v<c$). 
A non-overlapping average pooling is obtained when $v=0$. 
 
Based on the predefined matrices, 
the complete M-HOG can be formulated as follows:
\begin{equation} \label{eq.Bi}
f^{HOG}\left( {\bf{X}} \right) = {{\bf{Z}}_i^{HOG}}  = {{\bf{G}}_i} \circ {( {{\bf{L}}_{b,v}^T {{\bf{L}}_{b,v}{\bf{G}}_i^{\circ 2}{\bf{R}}_{b,v}}{\bf{R}}_{b,v}^T})^{{\circ\left({  - \frac{1}{2}}\right)}}},
\end{equation}
where 
the subscript $b$ is a  block size.
$ \circ $ is the Hadamard product.
${{\bf{G}}_i}$ is given by:
\begin{equation} \label{eq.Gi}
	{{\bf{G}}_i}{\bf{ = }}{\bf{L}}_{c_1,v}{{\bf{F}}_i}{\bf{R}}_{c_1,v}, \hspace{1cm} i = \{ 1,2,\dots,8\},
\end{equation} 
which is multiplied by the overlapped average pooling (${\bf{L}}_{c_1,v}$ and ${\bf{R}}_{{c_{1}},v}$) to improve the feature representation through the connection between neighbor blocks.
The gradient image ${{\bf{F}}_i}$ of HOG is given by:
\begin{equation} \label{eq.Fi}
	{{\bf{F}}_{i}^{} = {\bf{L}}_{c_2,0}({{\bf{Q}}_i} - {\bf{X}}}){\bf{R}}_{{c_{2}},0}, \hspace{1cm} i = \{ 1,2,\dots,8\},
\end{equation} 
where 
\begin{equation} \label{eq.PropSLBP}
	{\begin{array}{*{20}{l}}
			{{\bf{Q}}_{1}^{} = {\bf{V}}_1^{ - 1}{\bf{XH}}_1^{},}	&{{\bf{Q}}_{5}^{} = {\bf{V}}_1^{-2}{\bf{XH}}_1^{2},}\\
			{{\bf{Q}}_{2}^{} = {\bf{V}}_1^{ - 1}{\bf{X}},}		 	&{{\bf{Q}}_{6}^{} = {\bf{V}}_1^{-2}{\bf{X}},}\\
			{{{\bf{Q}}_{3}} = {\bf{V}}_1^{ - 1}{\bf{XH}}_1^{ - 1},}	&{{\bf{Q}}_{7}^{} = {\bf{V}}_1^{-2}{\bf{XH}}_1^{2},}\\
			{{\bf{Q}}_{4}^{} = {\bf{XH}}_1^{ - 1},}					&{{\bf{Q}}_{8}^{} = {\bf{XH}}_1^{-2}}.
	\end{array}}
\end{equation}
In equation \eqref{eq.PropSLBP}, ${\bf{Q}}_{1}$ to ${\bf{Q}}_{4}$ take the first-order gradients with one pixel distance difference and ${\bf{Q}}_{5}$ to ${\bf{Q}}_{8}$ include the second-order gradients with two pixel distance difference from the input $\bf{Q}$. 
$c_1$ and  $c_2$ are cell sizes for the overlapping and non-overlapping average pooling respectively.


In HOG \cite{dalal2005histograms}, the $L_2$-norm block normalization is defined as $\frac{{\bf{g}}}{{\sqrt {\left\| {\bf{g}} \right\|_2^2} }}$, where ${\bf{g}}$ is a vectorized histogram features in a given block.
For the matrix based block normalization with overlapping  in equation \eqref{eq.Bi}, ${{\bf{L}}_{b,v}{\bf{G}}_i^{\circ 2}{\bf{R}}_{b,v}}$ takes the sum of squares ${\left\| {\bf{g}} \right\|_2^2}$ for each block in a matrix-wise computation.
Next, the transposed projection matrices ${\bf{L}}_{b,v}^T$ and ${\bf{R}}_{b,v}^T$ are adopted to obtain the same dimension with ${\bf{G}}_i$ from the downsampled output of ${{\bf{L}}_{b,v}{\bf{G}}_i^{\circ 2}{\bf{R}}_{b,v}}$ due to the overlapping projection matrices ${\bf{L}}_{b,v}$ and ${\bf{R}}_{b,v}$. 
The superscript ${\circ \left({  - \frac{1}{2}}\right)}$ is the elementwise inverse of the squared root $\frac{1}{{\sqrt {  \cdot  } }}$. 
Then, each block of ${( {{\bf{L}}_{b,v}^T {{\bf{L}}_{b,v}{\bf{G}}_i^{\circ 2}{\bf{R}}_{b,v}}{\bf{R}}_{b,v}^T})^{{\circ\left({  - \frac{1}{2}}\right)}}}$ 
indicates $\frac{{1}}{{\sqrt {\left\| {\bf{g}} \right\|_2^2} }}$, and 
each block of ${\bf{G}}_i$ is same with ${\bf{g}} $.
By the Hadamard product, equation \eqref{eq.Bi} has included the matrix based block normalization which has the same output with $\frac{{\bf{g}}}{{\sqrt {\left\| {\bf{g}} \right\|_2^2} }}$ in HOG.
 
\subsection{A Matrix based LBP-HOG Descriptor}
\label{ssec:PropLBPHOG}
Based on the proposed M-LBP and M-HOG, the proposed M-LBP-HOG in matrix form can be directly and simply written as follows: 
\begin{equation} \label{eq.Bi2}
\footnotesize
\begin{array}{l}
f^{{{\scriptstyle LBP\hfill\atop
			\scriptstyle HOG\hfill}}}\left( {\bf{X}} \right) =
{\bf{Z}}_i^{{{\scriptstyle LBP\hfill\atop
					\scriptstyle HOG\hfill}}
		} = f^{HOG}\left( f^{LBP}\left( {\bf{X}} \right) \right),  i = \{ 1,2,\dots,8\}.
\end{array}
\end{equation}
For the final M-LBP-HOG features, 
each  ${\bf{Z}}_i^{{\scriptstyle LBP\hfill\atop 	\scriptstyle HOG\hfill}}$
is concatenated together into one feature vector.  \vspace{-0.2cm}

\begin{figure*}[h!]
	\centering \vspace{-0.6cm}
	\subfigure[Comparison of test accuracies at $b=2$]{ 
		\includegraphics[width=0.33\textwidth]{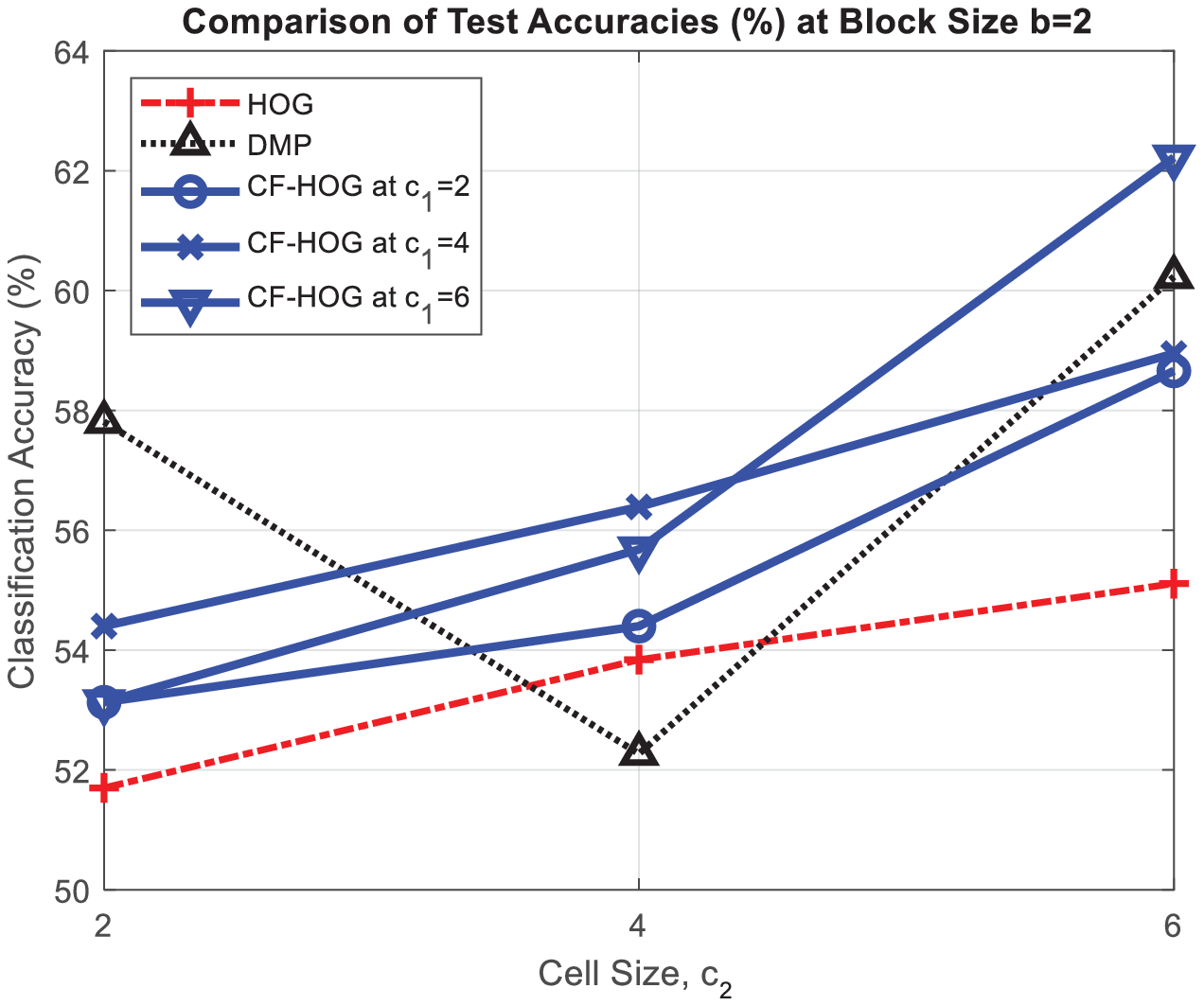}%
	} \hspace{-0.3cm}
	\subfigure[Comparison of test accuracies at $b=4$]{ 
		\includegraphics[width=0.33\textwidth]{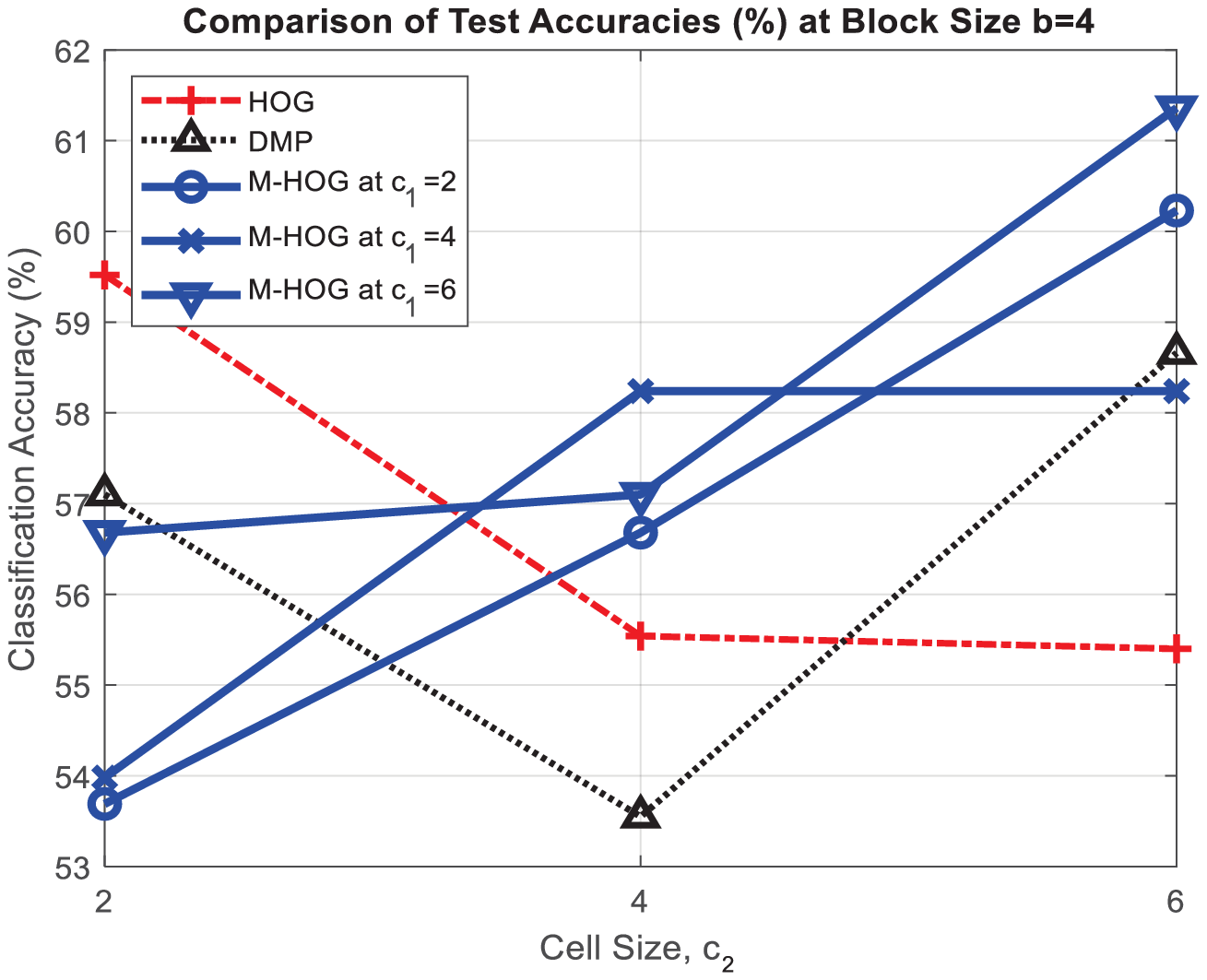}%
	} \hspace{-0.3cm}
	\subfigure[Comparison of test accuracies at $b=6$]{ 
		\includegraphics[width=0.33\textwidth]{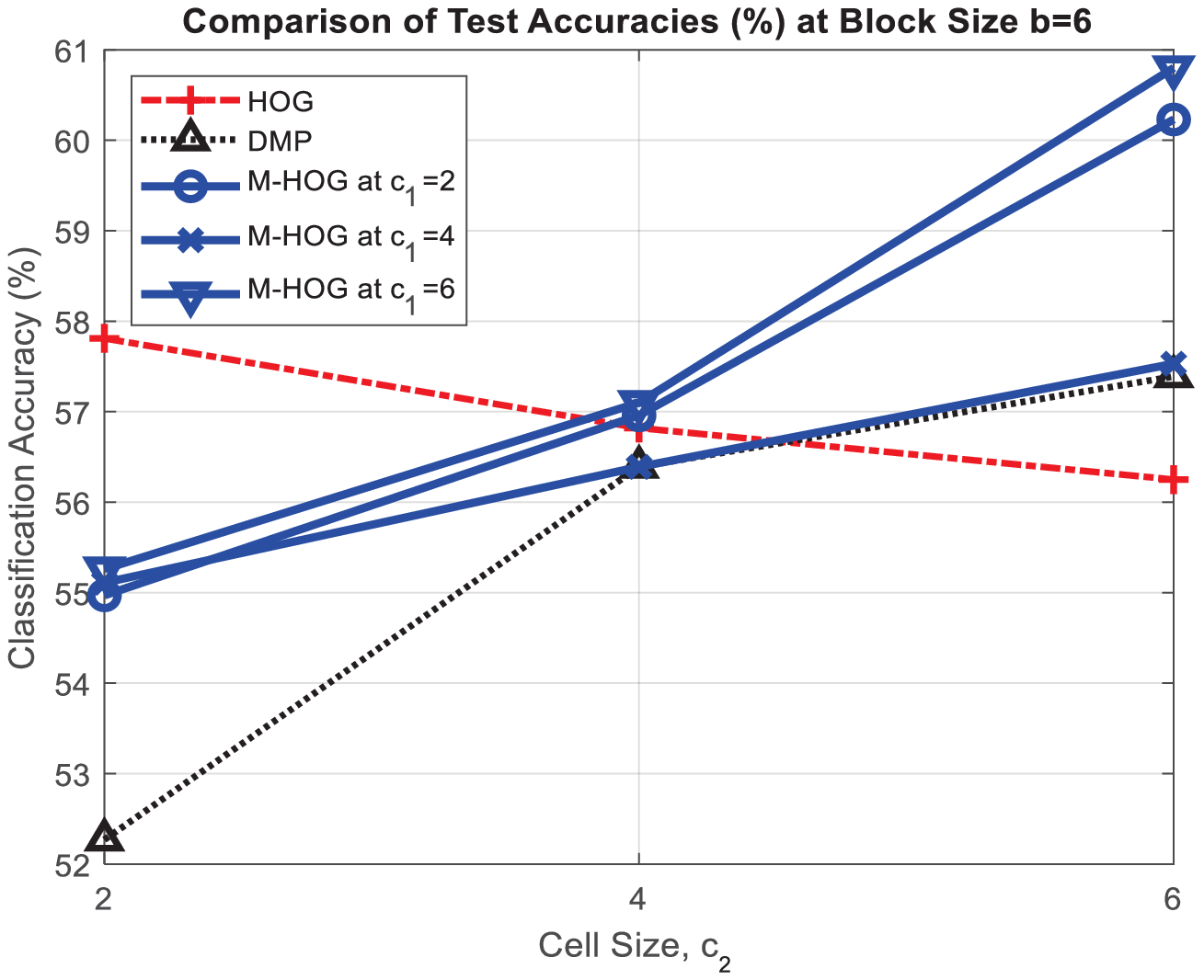}%
	} \vspace{-0.4cm}
	\caption{Comparison of HOG, DMP and M-HOG in terms of classification test accuracies (\%) on CBIS-DDSM database.}%
	\label{fig.HOGs}%
	\vspace{-0.4cm}
\end{figure*}

\section{Experiments}
\label{sec:exps}

In this section, we evaluate the proposed descriptors for mammogram classification. The experimental goals are as follows: 1) Observing the effect of the overlapping pooling among HOG, DMP and the proposed
M-HOG; 2) Performance comparison of the proposed M-LBP-HOG with state-of-the-arts methods. \vspace{-0.2cm} 
\subsection{Database and Experimental Setup}
\label{ssec:expData}

The most commonly used database in mammography is the Digital Database of Screening Mammography (DDSM) \cite{heath2000digital}. 
Recently, in \cite{lee2017curated}, a Curated Breast Imaging Subset of the DDSM (CBIS-DDSM) has been released with a standarized evaluation.
In this experiment, we used the CBIS-DDSM database.
The database includes $3,568$ ROI images which are categorized into two classes (malignant, benign) for $6,671$ patients and the images are resized to a $56 \times 56$ resolution. By following the data split setting in \cite{lee2017curated}, a training set ($2,864$ images) and a test set ($704$ images) are obtained.

For state-of-the-art descriptors,  LBP \cite{ojala1994performance},  HOG \cite{dalal2005histograms} and DMP \cite{liu2019pedestrian} are implemented for comparison. The LBP-HOG is also implemented based on a cascade of the LBP image and the HOG feature representation. 
For a state-of-the-art mammogram classification system, VGGNet \cite{xi2018abnormality} is also implemented.
For the parameter settings of each descriptor, we followed the general settings \cite{ojala1994performance}, \cite{dalal2005histograms}, \cite{liu2019pedestrian} for LBP, HOG, DMP, LBP-HOG, and for the proposed M-LBP, M-HOG and M-LBP-HOG: 
$P = 8$, 
${c_1}{\in}\left\{ {2,4,6} \right\}$, ${c_2}{\in}\left\{ {2,4,6} \right\}$, 
$v \in {\textstyle{c_2 \over 2}}$ 
and ${b}{\in}\left\{ {2,4,6} \right\}$. The histogram bin sizes of LBP and HOG are respectively fixed at $59$ and $9$ according to the parameter settings of LBP and HOG shown in \cite{ojala1994performance,dalal2005histograms}.
The LSE and SVM classifiers are utilized with a radial basis function (RBF). 
The regularization parameter of the LSE classifier is fixed at $0.0001$. 
According to \cite{xi2018abnormality}, the parameters settings of the VGGNet are set.
\vspace{-0.3cm}

\subsection{Observing the effect of the proposed overlapping between DMP and M-HOG}
\label{ssec:expCFhogvsDMP}
We show the effect of the proposed overlapping pooling of M-HOG 
since DMP only utilized the non-overlapping technique.
In order to observe the effectiveness, classification test accuracies were acquired based on the LSE classifier at different non-overlapping cell size ${c}_2$ and block size  $b$. 
Fig. \ref{fig.HOGs} shows the accuracy trends of the proposed overlapping pooling of  M-HOG at different overlapping sizes $v= \frac{c_1}{2}$ in accordance with the different cell sizes $c_1$, while ${c}_2$ and $b$ are fixed.
The accuracy performances are observed according to the increment of the non-overlapping pooling size $c_2$. According to the increment, the performance of  M-HOG is increasing while that of HOG and DMP are unstable.
The proposed M-HOG outperformed the other state-of-the-art methods when ${c}_2$ is $4$ and $6$. 
The best performance is observed in M-HOG where the overlapping pooling is included.  

    \begin{table}[t!]\centering
	\footnotesize\addtolength{\tabcolsep}{-3pt}
	\caption{Comparison of Classification Test Accuracies (\%)} 
	\label{tbl.acc}
		\begin{tabular}{|c|c|c|c|c|c|}
			\hline
		Method	& LBP    &  HOG  & DMP   &  LBP-HOG  & M-LBP-HOG  \\ \hline
		VGGNet	& \multicolumn{5}{c|}{63.35} \\ \hline
		SVM	&  60.79  &  60.80  &  60.51  &  63.49  &  62.36 \\ \hline
		LSE	& 56.25   &  59.52  &  60.23  &  62.07  &  \textbf{64.35} \\ \hline

	\end{tabular}  \vspace{-0.5cm}
\end{table}
\begin{table}[t!]\centering
	\scriptsize\addtolength{\tabcolsep}{-4.5pt}
	\caption{Comparison of CPU Processing Time in Seconds} 
	\label{tbl.cpu}
	\begin{tabular}{|l|r|c|c|c|}
		\hline
		{ }Method & \multicolumn{1}{c|}{\begin{tabular}[c]{@{}c@{}}Feature \\ Dimension\end{tabular}} & \multicolumn{1}{c|}{\begin{tabular}[c]{@{}c@{}}Feature  \\ Extraction Time\end{tabular}} & \multicolumn{1}{c|}{\begin{tabular}[c]{@{}c@{}}Training \\ Time \end{tabular}} & \multicolumn{1}{c|}{\begin{tabular}[c]{@{}c@{}}Test \\ Time \end{tabular}} \\ \hline
		{ }VGGNet    & 150,528             & -                 & 174,460       & { }209.88    \\ \hline
		{ }LBP        & 19,116             & 53.53                 & 24.40       & { }0.0069    \\ \hline
		{ }HOG        & 10,404             & 42.50                 & 14.59       & { }0.0038    \\ \hline
		{ }DMP        & 9,248              & 62.00                 & 12.97       & { }0.0034    \\ \hline
		{ }LBP-HOG    & 10,404             & 43.21                 & 14.25       & { }0.0038    \\ \hline
		{ }M-LBP-HOG & 1,800              & 31.13                 & 04.00        & { }0.0008    \\ \hline
	\end{tabular}  \vspace{-0.5cm}
\end{table} 

\subsection{Performance Comparison and Summary}
\label{ssec:expSummary}

In terms of classification performance, Table \ref{tbl.acc} shows the classification accuracies for the compared descriptors, namely LBP, HOG, DMP, LBP-HOG and the proposed M-LBP-HOG. The proposed M-LBP-HOG based on the LSE classifier shows {\bf the best performance} compared to the state-of-the-arts while LBP-HOG based on the SVM classifier and the VGGNet respectively  shows the second and the third bests.

In terms of computational performance, Table \ref{tbl.cpu} shows the CPU processing time in seconds. The averaged CPU times are reported over 10 runs. The proposed M-LBP-HOG based on the LSE classifier shows the best CPU time in the training and test phases due to the predefined projection matrices under global computation form. In addition, the M-LBP-HOG produces the smallest feature dimension which is 5 times less than the other descriptors.

As a summary, we showed 
1) the effectiveness of the overlapping pooling step in the proposed M-HOG compared with HOG and DMP;
2) In terms of classification accuracy, the proposed M-LBP-HOG with the LSE classifier achieved better performance with the smallest feature dimension than that of the other state-of-the-art methods; and 
3) In terms of CPU time, the proposed M-LBP-HOG achieved better performance than that of the other state-of-the-art methods due to the efficient computation using the global matrix multiplication.

\section{Conclusion}
\label{sec:conc}

Different from LBP and HOG involving the iterative scanning operations, we have presented the matrix based LBP (M-LBP) and HOG (M-HOG) using the matrix based pixel difference computation. In addition, the overlapping pooling in matrix form was also presented. The integrated formulation of M-LBP and M-HOG (M-LBP-HOG) was then proposed. The proposed descriptors were evaluated using the CBIS-DDSM database for mammogram classification. The M-LBP-HOG outperformed the state-of-the-art descriptors. 

\section*{\large Acknowledgment}
The authors are thankful to Prof. Hong-Goo Kang for his constructive comments and enormous support.

%


\bibliographystyle{IEEEbib}
\bibliography{refs}

\begin{thebibliography}{10}

\bibitem{bray2018global}
Freddie Bray, Jacques Ferlay, Isabelle Soerjomataram, Rebecca~L. Siegel,
  Lindsey~A. Torre, and Ahmedin Jemal,
\newblock ``Global cancer statistics 2018: Globocan estimates of incidence and
  mortality worldwide for 36 cancers in 185 countries,''
\newblock {\em CA: A Cancer Journal For Clinicians}, vol. 68, no. 6, pp.
  394--424, 2018.

\bibitem{soltanian2001shape}
Hamid Soltanian-Zadeh, Siamak Pourabdollah-Nezhad, and Farshid~Rafiee Rad,
\newblock ``Shape-based and texture-based feature extraction for classification
  of microcalcifications in mammograms,''
\newblock in {\em Medical Imaging: Image Processing}. International Society for
  Optics and Photonics, 2001, vol. 4322, pp. 301--311.

\bibitem{bekker2016multi}
Alan~Joseph Bekker, Moran Shalhon, Hayit Greenspan, and Jacob Goldberger,
\newblock ``Multi-view probabilistic classification of breast
  microcalcifications,''
\newblock {\em IEEE Transactions on Medical Imaging}, vol. 35, no. 2, pp.
  645--653, 2016.

\bibitem{oliver2007classifying}
Arnau Oliver, Xavier Llad{\'o}, Robert Marti, Jordi Freixenet, and Reyer
  Zwiggelaar,
\newblock ``Classifying mammograms using texture information,''
\newblock in {\em Medical Image Understanding and Analysis}. Citeseer, 2007,
  vol. 223.

\bibitem{shastri2018density}
Aditya~A. Shastri, Deepti Tamrakar, and Kapil Ahuja,
\newblock ``Density-wise two stage mammogram classification using texture
  exploiting descriptors,''
\newblock {\em Expert Systems with Applications}, vol. 99, pp. 71--82, 2018.

\bibitem{ojala1994performance}
Timo Ojala, Matti Pietikainen, and David Harwood,
\newblock ``Performance evaluation of texture measures with classification
  based on kullback discrimination of distributions,''
\newblock in {\em International Conference on Pattern Recognition (ICPR)}.
  IEEE, 1994, vol.~1, pp. 582--585.

\bibitem{dalal2005histograms}
Navneet Dalal and Bill Triggs,
\newblock ``Histograms of oriented gradients for human detection,''
\newblock in {\em IEEE Conference on Computer Vision and Pattern Recognition
  (CVPR)}. IEEE, 2005, vol.~1, pp. 886--893.

\bibitem{liu2017local}
Li~Liu, Paul Fieguth, Yulan Guo, Xiaogang Wang, and Matti Pietik{\"a}inen,
\newblock ``Local binary features for texture classification: Taxonomy and
  experimental study,''
\newblock {\em Pattern Recognition}, vol. 62, pp. 135--160, 2017.

\bibitem{wang2009hog}
Xiaoyu Wang, Tony~X Han, and Shuicheng Yan,
\newblock ``An hog-lbp human detector with partial occlusion handling,''
\newblock in {\em 2009 IEEE 12th international conference on computer vision}.
  IEEE, 2009, pp. 32--39.

\bibitem{konstantinidis2016building}
Dimitrios Konstantinidis, Tania Stathaki, Vasileios Argyriou, and Nikolaos
  Grammalidis,
\newblock ``Building detection using enhanced hog--lbp features and region
  refinement processes,''
\newblock {\em IEEE Journal of Selected Topics in Applied Earth Observations
  and Remote Sensing}, vol. 10, no. 3, pp. 888--905, 2016.

\bibitem{zhang2011boosted}
Junge Zhang, Kaiqi Huang, Yinan Yu, Tieniu Tan, et~al.,
\newblock ``Boosted local structured hog-lbp for object localization,''
\newblock 2011.

\bibitem{liu2019pedestrian}
Xing Liu, Kar-Ann Toh, and Jan~P Allebach,
\newblock ``Pedestrian detection using pixel difference matrix projection,''
\newblock {\em IEEE Transactions on Intelligent Transportation Systems}, 2019.

\bibitem{ahonen2006face}
Timo Ahonen, Abdenour Hadid, and Matti Pietikainen,
\newblock ``Face description with local binary patterns: Application to face
  recognition,''
\newblock {\em IEEE Transactions on Pattern Analysis and Machine Intelligence},
  vol. 28, no. 12, pp. 2037--2041, 2006.

\bibitem{sepas2017light}
Alireza Sepas-Moghaddam, Paulo~Lobato Correia, and Fernando Pereira,
\newblock ``Light field local binary patterns description for face
  recognition,''
\newblock in {\em IEEE International Conference on Image Processing (ICIP)}.
  IEEE, 2017, pp. 3815--3819.

\bibitem{han2009palmprint}
Yufei Han, Zhenan Sun, and Tieniu Tan,
\newblock ``Palmprint recognition using coarse-to-fine statistical image
  representation,''
\newblock in {\em IEEE International Conference on Image Processing (ICIP)}.
  IEEE, 2009, pp. 1969--1972.

\bibitem{heath2000digital}
Michael Heath, Kevin Bowyer, Daniel Kopans, Richard Moore, and W.~Philip
  Kegelmeyer,
\newblock ``The digital database for screening mammography,''
\newblock in {\em International Workshop on Digital Mammography}. Medical
  Physics Publishing, 2000, pp. 212--218.

\bibitem{lee2017curated}
Rebecca~Sawyer Lee, Francisco Gimenez, Assaf Hoogi, Kanae~Kawai Miyake, Mia
  Gorovoy, and Daniel~L. Rubin,
\newblock ``A curated mammography data set for use in computer-aided detection
  and diagnosis research,''
\newblock {\em Scientific Data}, vol. 4, pp. 170--177, 2017.

\bibitem{xi2018abnormality}
Pengcheng Xi, Chang Shu, and Rafik Goubran,
\newblock ``Abnormality detection in mammography using deep convolutional
  neural networks,''
\newblock in {\em IEEE International Symposium on Medical Measurements and
  Applications (MeMeA)}. IEEE, 2018.

\end{thebibliography}

\end{document}